\documentclass[journal]{IEEEtran}

\usepackage{hyperref}
\usepackage{framed,multirow}
\usepackage{cite}
\usepackage{amsmath, amssymb}
\usepackage{amsthm}
\usepackage{graphicx}
\usepackage{amsfonts}
\usepackage{longtable}
\usepackage{epsf, epsfig}
\usepackage{epstopdf}
\usepackage{subfigure}
\usepackage{algorithm}
\usepackage{algorithmic}
\usepackage{color}
\usepackage{booktabs}
\usepackage{adjustbox}
\usepackage{xspace}

\usepackage{xargs} 
\usepackage[pdftex,dvipsnames]{xcolor}



\graphicspath{{images/}}

\newcommand{\ie}{\emph{i.e.}\xspace}
\newcommand{\eg}{\emph{e.g.}\xspace}
\newcommand{\etal}{\emph{et al.}\xspace}

\newcommand{\vct}[1]{\boldsymbol{#1}} 
\newcommand{\mat}[1]{\boldsymbol{#1}} 

\newcommand{\T}{^{\textrm T}} 

\DeclareMathOperator*{\argmin}{arg\,min}

\newcommand{\eat}[1]{}

\begin{document}

\title{Convolutional Ordinal Regression Forest for Image Ordinal Estimation}

\author{\normalsize{
Haiping Zhu,
Hongming Shan,~\IEEEmembership{Member, IEEE},
Yuheng Zhang, 
Lingfu Che,
Xiaoyang Xu,
Junping Zhang,~\IEEEmembership{Member, IEEE}, 
Jianbo Shi,
Fei-Yue Wang,~\IEEEmembership{Fellow, IEEE}

\thanks{\textcopyright  2021 IEEE.  Personal use of this material is permitted.  Permission from IEEE must be obtained for all other uses, in any current or future media, including reprinting/republishing this material for advertising or promotional purposes, creating new collective works, for resale or redistribution to servers or lists, or reuse of any copyrighted component of this work in other works.}
\thanks{This work was supported in part by the National Key R \& D Program of China (No. 2018YFB1305104), the Shanghai Municipal Science and Technology Major Project (No. 2018SHZDZX01) and ZJLab, National Natural Science Foundation of China (NSFC 61673118) , the Shanghai Center for Brain Science and Brain-inspired Technology, National Natural Science Foundation of China (NSFC 61533019), National Natural Science Foundation of China (U1811463), Key-Area Research and Development Program of Guangdong Province (No. 2020B090921003), and Intel Collaborative Research Institute for Intelligent and Automated Connected Vehicles (``ICRI-IACV'').
\emph{(Haiping Zhu and Hongming Shan are co-first authors.)} \emph{(Corresponding authors: Junping Zhang; Fei-Yue Wang.)}}
\thanks{H. Zhu, Y. Zhang, L. Che, X. Xu, and J. Zhang are with the Shanghai Key Laboratory of Intelligent Information Processing, School of Computer Science, Fudan
University, Shanghai 200433, China (e-mail: hpzhu14@fudan.edu.cn; yuhengzhang16@fudan.edu.cn;  lfche16@fudan.edu.cn; xuxy17@fudan.edu.cn; jpzhang@fudan.edu.cn).}
\thanks{H. Shan is with the Institute of Science and Technology for Brain-inspired Intelligence and  MOE Frontiers Center for Brain Science, Fudan University, Shanghai, 200433, China, and also with the Shanghai Center for Brain Science and Brain-inspired Technology, Shanghai 201210, China (e-mail: hmshan@fudan.edu.cn).}
\thanks{J. Shi is with the GRASP Laboratory, University of Pennsylvania, Philadelphia, PA USA (e-mail: jshi@seas.upenn.edu). }
\thanks{F.Y. Wang is with the State Key Laboratory for Management and Control of Complex Systems, Institute of Automation, Chinese Academy of Sciences, Beijing 100190, China, and also with the Institute of Systems Engineering, Macau University of Science and Technology, Macau 999078, China, and also with the University of Chinese Academy of Sciences, Beijing 100049, China (e-mail: feiyue@ieee.org). }
}

}

\markboth{IEEE TRANSACTIONS ON NEURAL NETWORKS AND LEARNING SYSTEMS, VOL. XX, NO. XX, XXX 2020}%
{Shell \MakeLowercase{\textit{et al.}}: Bare Demo of IEEEtran.cls for Journals}

\maketitle

\begin{abstract}
Image ordinal estimation is to predict the ordinal label of a given image, which can be categorized as an ordinal regression problem.
Recent methods formulate an ordinal regression problem as a series of binary classification problems. 
Such methods cannot ensure that the global ordinal relationship is preserved since the relationships among different binary classifiers are neglected. 
We propose a novel ordinal regression approach, termed Convolutional Ordinal Regression Forest or CORF, for image ordinal estimation, which can integrate ordinal regression and differentiable decision trees with a convolutional neural network for obtaining precise and stable global ordinal relationships. 
The advantages of the proposed CORF are twofold. First, instead of learning a series of binary classifiers \emph{independently}, the proposed method aims at learning an ordinal distribution for ordinal regression by optimizing those binary classifiers \emph{simultaneously}. Second, the differentiable decision trees in the proposed CORF can be trained together with the ordinal distribution in an end-to-end manner. The effectiveness of the proposed CORF is verified on two image ordinal estimation tasks, \ie facial age estimation and image aesthetic assessment, showing significant improvements and better stability over the state-of-the-art ordinal regression methods.
\end{abstract}


\section{Introduction}

Image ordinal estimation is to infer the ordinal label of a given image, which can be used to address many real-world problems such as disease rating~\cite{bender1997ordinal,doyle2014predicting,perez2011ordinal,cardoso2005modelling,perez2014organ}, age estimation~\cite{niu2016ordinal,chen2017using}, image aesthetic assessment~\cite{luo2011content,kong2016photo}, etc. This task can be 
categorized as an ordinal regression problem~\cite{harrell2015regression,herbrich1999support}, an intermediate problem between classification and regression.
Imagining a scenario where a doctor is asked to make an assessment of the severity of a patient's illness. The assessment scale can be \{\emph{very healthy, healthy, slightly sick, sick, serious illness}\}. 
This is not a simple classification task, because the answer is not a clear right or wrong answer. But it is also not a continuous regression task, because we only care about the discrete sorting between different answers.

\begin{figure*}[!t]
\centering
\setlength\abovecaptionskip{-0.7\baselineskip}
\includegraphics[width=1\linewidth]{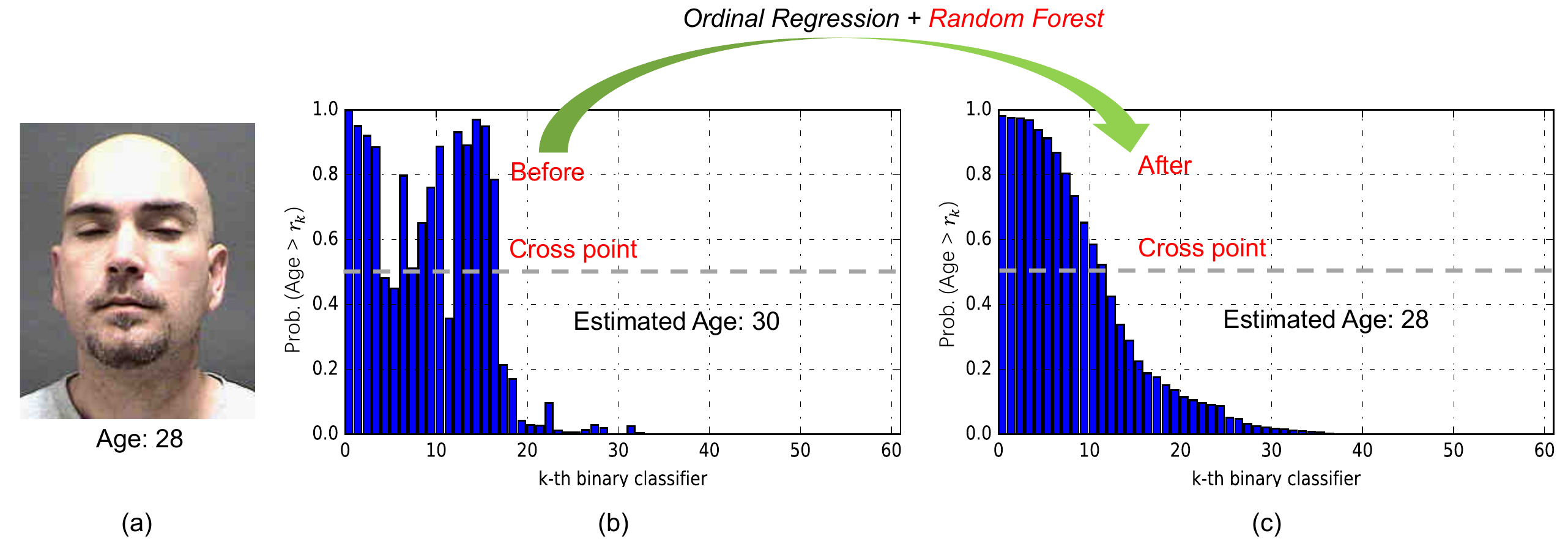} 
\caption{Ordinal regression on the facial age estimation. Each $k$-th classifier computes whether the probability of one person’s age is greater than $r_k$, where $r_k= r_1 + k - 1$ and $r_1 = 16$. The cross point of probability $> 0.5$ determines the person’s age. (a) The input face of age 28. (b) Current methods fail to preserve the global ordinal relationship among binary classifiers. (c) The CORF produces ordinal distribution with well-preserved global relationships.}
\label{toy_example}
\end{figure*}

To solve the ordinal regression problem, traditional methods formulate it as a classification problem, which can be solved by many well-studied classification algorithms~\cite{shashua2003ranking,crammer2002pranking,chu2007support,lin2006large,Riccardi2014Cost,PerezProjection,Gu2015Incremental,XiaoMultiple}. However, the ordinal information is not used to improve the predictive performance. Recently,  researchers formulate it as a series of binary classifiers to model ordinal relationship~\cite{niu2016ordinal,chen2017using,frank2001simple,lin2012reduction,fu2018deep}---each binary classifier recognizes a sample into two classes by judging whether its label is less than a specified class attribute. With more attention given to the ordinal information, these binary classification methods lead to better predictive performance. However, these methods learn a series of binary classifiers independently, thus the underlying global relationship among these binary classifiers is neglected. Fig.~\ref{toy_example} (b) presents an example of such a deficiency, where the output probability should be a decreasing trend, rather than increasing sometimes.

To overcome this deficiency that lacks global ordinal relationships, we present a convolutional ordinal regression forest (CORF) approach for the image ordinal estimation. Our method utilizes the differentiable decision trees~\cite{kontschieder2015deep} to preserve the ordinal relationship by learning an ordinal distribution rather than learning a series of binary classifiers independently. 
There are three advantages of using the differentiable decision tree to handle the ordinal regression tasks. First, the decision tree has the expressive power to model any ordinal distribution through a mixture of leaf node predictions. Second, one can learn the parameters of the split nodes in the differentiable decision trees by optimizing tree learning in combination with deep learning in an end-to-end manner. Third, the ensemble property of a random forest can alleviate the over-fitting issue. More specifically, we regard the output of a decision tree as an ordinal distribution, followed by optimizing the CORF with the cross-entropy loss between the predicted and ground-truth ordinal distribution. We optimize the trees by updating split nodes and leaf nodes alternatively. 
Following~\cite{shen2018deep}, the leaf node predictions can be optimized by variational boundary~\cite{jordan1999introduction,yuille2003concave} after fixing the split node, where its upper bound can be used as the alternative loss function to be minimized. Next, we employ the averaged loss of all decision trees in a forest as the overall loss function to learn a forest and make the split nodes of different trees connect to the same feature space from deep neural network. As a result, the split node parameters of all decision trees in a forest can be learned simultaneously from the training data.

The contributions of this paper are summarized as follows.
\begin{enumerate}
	\item We present a convolutional ordinal regression forest (CORF) approach to address the ordinal regression problem, in which the ordinal relationship can be well preserved. This is the first work to integrate ordinal regression and differentiable decision tree with a convolutional neural network for image ordinal estimation tasks.
	\item We provide a detailed derivation for the update rule of the leaf nodes.
	\item Experimental results on five benchmark datasets demonstrate the effectiveness and the stability of our CORF in modeling the ordinal distribution for two image ordinal estimation tasks---facial age estimation and image aesthetic assessment.
\end{enumerate}

The rest of this paper is organized as follows. We first briefly survey the development of the random forest methods, with emphasis on the differentiable decision tree, and the ordinal regression methods in Sec.~\ref{related}.  We then detail the proposed method in Sec.~\ref{method}. Finally, we analyze our experimental results in Sec.~\ref{exp}, followed by a concluding summary in Sec.~\ref{conc}.

\section{Related Work}\label{related}

Since the CORF extends the differentiable decision trees to ordinal regression problems, this section briefly surveys the developments of random forests and ordinal regression.

\subsection{Random Forest} 

Random forest is an ensemble of decision trees~\cite{amit1997shape,breiman2001random,criminisi2013decision,ho1995random}, each of which usually consists of several split nodes and leaf nodes and is built on a randomly selected subset of samples and features. These decision trees in a random forest are constructed independently. For the regression problems, the output of a random forest is the average of the outputs of all decision trees, while for the classification problems, a voting scheme is typically used to determine the final outputs based on all the outputs of leaf nodes. It should be noted that the decision trees are the hierarchical mixture of expert models~\cite{jordan1994hierarchical,bishop2002bayesian}, which are explainable and can learn the hierarchical structure based on pre-specified features with data-driven architectures. 

Recently, an increasing trend is to combine decision trees and deep neural networks in a unified framework for powerful representative learning. Since the construction of traditional decision trees usually employs heuristic algorithms such as greedy algorithms to implement a hard partition decision for each split node~\cite{amit1997shape,liu2019embedded}, it is intractable to construct decision trees and learn deep neural network simultaneously trough the back-propagation algorithm.
Some effort has been made to integrate these two different worlds~\cite{kontschieder2015deep,roy2016monocular,shen2017label,shen2018deep,tanno2019adaptive}. Kontschieder~\etal~\cite{kontschieder2015deep} proposed a \emph{differentiable decision tree} to combine these two worlds by defining a soft partition function for each split node, which ensures that the decision trees are differentiable and enables the decision trees to be jointly learned with deep neural networks. Next, Shen~\etal~\cite{shen2017label,shen2018deep} extended this differentiable decision tree to solve the problems of label distribution learning and regression learning. Similarly, the neural regression forest (NRF)~\cite{roy2016monocular} was proposed for depth estimation by using a tree of CNNs to jointly learn the feature representation and decision functions for each node in the tree. The recently proposed adaptive neural trees (ANTs)~\cite{tanno2019adaptive} incorporate representation learning into edges, routing functions, and leaf nodes of a decision tree, along with a back-propagation-based training algorithm that adaptively grows the architecture from primitive modules such as convolutional layer. In this paper, our method utilizes this differentiable decision tree proposed by dNDFs~\cite{kontschieder2015deep} for image ordinal estimation, which has a significant difference in the leaf node distribution---ordinal distribution.

\begin{figure*}[!t]
\setlength\abovecaptionskip{-0.7\baselineskip}
\centering
\includegraphics[width=1\linewidth]{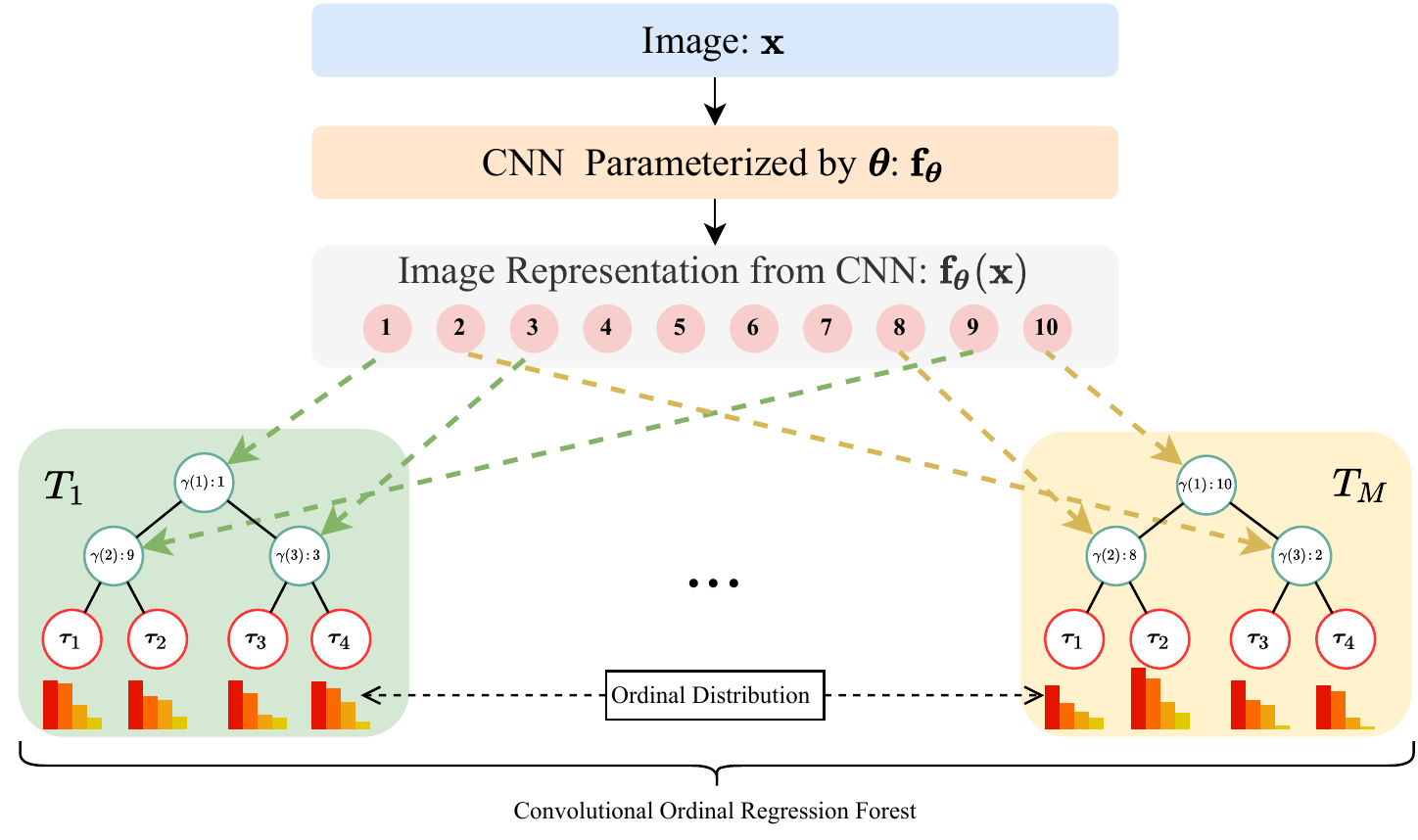}
\caption{The illustration of our CORF. An input image $\vct{x}$ is represented by a vector $\vct{f}_{\vct\theta}$ that is extracted from a convolutional neural network (CNN) parameterized by $\vct\theta$. 
The teal and red circles are split nodes and leaf nodes, respectively. Index function $\gamma$ is used to assign features to different nodes. The dashed arrows indicate the correspondence between the split nodes of the two trees and the image representation of the input image. Each tree has independent leaf node ordinal distribution, and the final output of a forest is the average output of all these $M$ trees.}
\label{network}
\end{figure*}

\subsection{Ordinal Regression}

Ordinal regression aims to learn a rule to predict ordinal labels. Traditional ordinal regression algorithms are modified from well-known classification methods~\cite{albert1997bayesian,herbrich1999support,shashua2003ranking,chu2005gaussian,crammer2002pranking,chu2007support,lin2006large,dembczynski2007ordinal,agresti2010analysis,Gu2015Incremental,XiaoMultiple}. For instances, Herbrich~\etal~\cite{herbrich1999support} utilized support vector machine (SVM) for ordinal regression, and then Shashua~\etal~\cite{shashua2003ranking} refined SVM to handle multiple thresholds. Crammer~\etal~\cite{crammer2002pranking} proposed the perceptron ranking algorithm to generalize the online perceptron algorithm with multiple thresholds for ordinal regression. Lin~\etal~\cite{lin2006large} proposed a thresholded ensemble model for ordinal regression problems, which consists of a weighted ensemble of confidence functions and an ordered vector of thresholds. However, these methods formulate the ordinal regression as a classification task, ignoring the potential of ordinal information in improving predictive performance. Recently, some methods formulate the ordinal regression as a series of binary classifications~\cite{niu2016ordinal,chen2017using,frank2001simple,fu2018deep,li2007ordinal,lin2012reduction}. For examples, Frank~\etal~\cite{frank2001simple} utilized decision trees as binary classifiers for ordinal regression. Li~\etal~\cite{li2007ordinal} proposed a framework to reduce an ordinal regression problem as a set of classification problems, and employed an SVM to solve the classification problems. Niu~\etal~\cite{niu2016ordinal} introduced a CNN network with multiple binary outputs to solve the ordinal regression for age estimation. Chen~\etal~\cite{chen2017using} proposed to learn multiple binary CNNs, and then aggregated the final outputs. With more attention given to the ordinal information, these binary classification methods lead to better predictive performance. However, these methods learn each binary classifier separately and independently, ignoring the global ordinal relationship among these binary classifiers. To better preserve the ordinal relationship among these binary classifiers, in this paper, we propose a novel ordinal regression method by extending the differentiable decision trees to deal with the ordinal regression problems.

\section{Convolutional Ordinal Regression Forest}\label{method}

In this section, we first formulate the problem of image ordinal estimation in Sec.~\ref{sec_ordinal_reg}. Next, we present the ordinal regression tree involving how to construct a decision tree and the objective function in Sec.~\ref{sec_decision_tree}, which can form a convolutional ordinal regression forest in Sec.~\ref{sec_CORF}. Last, Sec.~\ref{learning_split_node} describes how to optimize a single ordinal regression tree with convolutional neural networks and, which can then be extended to a forest.

\subsection{Image Ordinal Estimation}\label{sec_ordinal_reg}
Let $\mathcal{X}$ be the image space and $\mathcal{Y}=\{r_1, r_2, \ldots, r_K\}$ be the label space with ordered ranks $r_1 \preceq r_2 \preceq \ldots \preceq r_K $, where $K$ is the total number of possible label ranks and the symbol $\preceq$ denotes the order among different ranks. The image ordinal estimation is to learn a mapping function from the input image to its ordinal label, saying $h:\mathcal{X} \to \mathcal{Y}$. Following recent studies~\cite{niu2016ordinal,chen2017using,frank2001simple,fu2018deep}, we divide the original ordinal regression problem with $K$ ranks into $K-1$ binary classification tasks. For each rank $r_k$, $k=1, \ldots, K-1$, a binary classifier $g_k$ is constructed to classify an image $\vct{x}\in\mathcal{X}$ into two classes, and the ground-truth label of the image $\vct{x}$ depends on whether its ordinal label $y$ is greater than $r_k$; the label of $\vct{x}$ for classifier $g_k$ is $1$ if $y > r_k$ and $0$ otherwise. Note that the output of the $k$-th classifier, $g_k$, is a probability between $0$ and $1$. Thus, the final ordinal label of an image $\vct{x}$ can be predicted as
\begin{equation}\label{cal_final_output}
h(\vct{x}) = r_1 + \eta\sum_{k=1}^{K-1}\mat{1}\left[g_k(\vct{x}) > 0.5\right],
\end{equation}
where $\eta$ is the partitioning interval, $\mat{1}[\cdot]$ denotes the 0-1 indicator function; the output is 1 if the inner condition holds, and 0 otherwise. 

Let $\mathcal{D}\in\mathbb{R}^{K-1}$ denote an ordinal distribution label space, we can transform the ordinal label $y$ as an ordinal distribution label $\vct{d} = (d^1, d^2, \ldots, d^{K-1})\T\in\mathcal{D}$, where $d^k = 1$ if $y > r_k$, and $d^k = 0$ otherwise. 
Therefore, the goal of an image ordinal estimation task is instead to learn a mapping function $\vct{g}:\mathcal{X} \to \mathcal{D}$, where the binary classifier $g_k$ is the $k$-th element in $\vct{g}$.

\subsection{Ordinal Regression Tree}\label{sec_decision_tree}

There are several ways to learn such a mapping function $\vct{g}$ from the image space to ordinal label space, such as OR-CNN~\cite{niu2016ordinal} and Ranking-CNN~\cite{chen2017using}. Inspired by the promising results of differentiable decision trees\cite{kontschieder2015deep,shen2018deep}, we apply the differentiable decision trees to learn the ordinal distribution, termed ordinal regression tree. Following the definition of the image ordinal estimation in Sec.~\ref{sec_ordinal_reg}, our goal is to learn a mapping function $\vct{g}$ through the ordinal regression tree that can take the output of the convolutional neural network as the input.

\subsubsection{Constructing Decision Trees}
We start by introducing the decision tree first.
Let $T$ denote a decision tree. An input image is represented by a vector $\vct{f}_{\vct{\theta}}$ extracted from convolutional neural networks (CNNs) such as VGG-16~\cite{simonyan2014very} and ResNet-50~\cite{he2016deep}. Each decision tree $T$ is defined as $(\mathcal{N}, \mathcal{L})$, where $\mathcal{N}$ represents a set of split nodes to decide which child node a sample should be assigned to, and $\mathcal{L}$ denotes a set of leaf nodes to output the probability. 
More specifically, we define a decision function, $s_n: \mathcal{X}\to\{0, 1\}$ for $n$-th split node, parameterized by a selected feature in $\vct{f}_{\vct{\theta}}$ to determine which child node (left or right) a sample should be assigned to. We also assume that each leaf node $l\in\mathcal{L}$ outputs an ordinal distribution $\vct{\tau}_l = (\tau_{l}^1, \tau_{l}^2, \cdots, \tau_{l}^{K-1})\T$ over $\mathcal{D}$

Following~\cite{kontschieder2015deep,shen2017label}, we also use the decision function $s_n(\vct{f}_{\vct{\theta}}(\vct{x})) = \sigma(\vct{f}_{\vct{\theta}}(\vct{x})_{\gamma(n)})$ to build a differentiable decision tree, where  $\sigma(\cdot)$ is a sigmoid function and  $\gamma(\cdot)$ is an index function that specify the $\gamma(n)$-th element of $\vct{f}(\vct{x})$ for the corresponding split node $n$.
In principle, the function $\vct{f}_{\vct{\theta}}$ can be any function. 
In this paper, we use a convolutional neural network as the feature function $\vct{f}_{\vct{\theta}}$, which can be learned in an end-to-end manner. The index function $\gamma(\cdot)$ between the split nodes and the output units of function $\vct{f}$ is initialized randomly before the tree learning. An example shown in Fig.~\ref{network} demonstrates how to construct a convolutional ordinal regression forest containing $M$ trees for learning ordinal distribution. Then, the probability of a sample $\vct{x}$ falling into leaf node $l$ is given by
\begin{align}
p(l|\vct{x};\vct{\theta}) = \prod_{n\in\mathcal{N}} &s_n(\vct{f}_{\vct{\theta}}(\vct{x}))^{\vct{1}\left[l\in\mathcal{L}_n^l\right]}\notag \\
&\times\left(1 - s_n(\vct{f}_{\vct{\theta}}(\vct{x}))\right)^{\vct{1}\left[l\in\mathcal{L}_n^r\right]},
\end{align}
where $\vct{1}[\cdot]$ is the 0-1 indicator function, and $\mathcal{L}_n^l$ and $\mathcal{L}_n^r$ denote the sets of leaf nodes held by the left and right subtrees of node $n$, $T_n^l$ and $T_n^r$, respectively. Therefore, given a sample $\vct{x}$, the mapping function $\vct{g}$, based on the decision tree $T$, is defined by
\begin{equation}
\vct{g}(\vct{x}; \vct{\theta}, T) = \sum_{l\in\mathcal{L}} p(l|\vct{x}; \vct{\theta})\vct{\tau}_l.
\end{equation}

\subsubsection{Objective Function}\label{optimize}
Given a training set of $N$ data pairs, $\mathcal{S}=\{(\vct{x}_i, \vct{d}_i)\}_{i=1}^N$, our goal is to learn a decision tree $T$ described in Sec.~\ref{sec_ordinal_reg}, which can output an ordinal distribution $\vct{g}(\vct{x}; \vct{\theta}, T)$ in reference to $\vct{d}$ for an input image $\vct{x}$. A straightforward way to measure the difference between the estimated  and ground-truth ordinal distribution is the cross entropy loss between $\vct{g}(\vct{x}; \vct{\theta}, T)$ and $\vct{d}$, therefore, the overall objective function over $\mathcal{S}$ can be written as
\begin{align}
\label{eq_obj}
&\quad\ell_{\mathcal{S}}(\vct{\theta}, \vct{\tau}) \\
&= -\frac{1}{N}\sum_{i=1}^{N}\sum_{k=1}^{K-1}\Big(d_i^k\log\big({g_k(\vct{x}_i;\vct{\theta}, T)\big)} \notag\\ 
& \quad\quad\quad\quad\quad\quad\quad\quad + (1 - d_i^k)\log\big(1 - {g_k(\vct{x}_i;\vct{\theta}, T)\big)}\Big) \notag\\
&= -\frac{1}{N}\sum_{i=1}^{N}\sum_{k=1}^{K-1}\Big(d_i^k\log\big(\sum_{l\in\mathcal{L}}p(l|\vct{x}_i;\vct{\theta})\tau_{l}^k\big) \notag \\
&\quad\quad\quad\quad\quad\quad\quad\quad +(1 - d_i^k)\log\big(1 - \sum_{l\in\mathcal{L}}p(l|\vct{x}_i;\vct{\theta})\tau_l^k\big)\Big), \notag \\
&= -\frac{1}{N}\sum_{i=1}^{N}\sum_{k=1}^{K-1}\Big(d_i^k\log\big(\sum_{l\in\mathcal{L}}p(l|\vct{x}_i;\vct{\theta})\tau_{l}^k\big) \notag \\
&\quad\quad\quad\quad\quad\quad\quad\quad+(1 - d_i^k)\log\big(\sum_{l\in\mathcal{L}}p(l|\vct{x}_i;\vct{\theta})(1 - \tau_l^k)\big)\Big).\notag
\end{align}
For the last equation, $\sum_{l\in\mathcal{L}}p(l|\vct{x}_i;\vct{\theta})=1$ is used. To simplify the math symbols, we rewrite $d_i^k$ and $(1-d_i^k)$ as $d_i^{(k, c=1)}$ and $d_i^{(k,c=2)}$, respectively. Similarly, we rewrite $\tau_{l}^k$ and $(1 - \tau_{l}^k)$ as $\tau_l^{(k,c=1)}$ and $\tau_l^{(k,c=2)}$, respectively. Clearly, $\sum_{c=1}^2d_i^{(k, c)}=1$ and $\sum_{c=1}^2\tau_l^{(k,c)} = 1$. The objective function in Eq.~\eqref{eq_obj} can then be simplified as:
\begin{align}\label{eq_simple_obj}
\ell_{\mathcal{S}}(\vct{\theta}, \vct{\tau}) \!
=\! -\frac{1}{N}&\!\sum_{i=1}^{N}\!\sum_{k=1}^{K-1}\sum_{c=1}^{2}\bigg(d_i^{(k,c)}\log\big(\!\sum_{l\in\mathcal{L}}p(l|\vct{x}_i;\vct{\theta})\tau_{l}^{(k,c)}\big)\bigg) \notag \\
&\quad\quad\quad\quad \mathrm{s.t.}\ \ \forall\ l, k,\ \  \sum_{c=1}^{2}\tau_{l}^{(k,c)} = 1.
\end{align}

\subsection{Convolutional Ordinal Regression Forest}\label{sec_CORF}

We denote a collection of ordinal regression trees with a convolutional neural network as a convolutional ordinal regression forest (CORF) $\mathcal{F}=\{T_1, T_2, \ldots, T_M\}$, where $T_i$ represents the $i$-th ordinal regression tree. In the training phase, all decision trees in a forest $\mathcal{F}$ are based on the same feature $\vct{f}_{\vct\theta}(\cdot)$, but with different index functions $\gamma$ in decision trees so that the ordinal distributions in left nodes are independent. Since a forest is a set of decision trees, the overall loss function for a forest can be defined as the average of loss functions of all trees. That is,
\begin{align}
\ell^\mathcal{F} = \frac{1}{M}\sum_{m=1}^{M}\ell^{{T}_m}, 
\end{align}
where $\ell^{{T}_m}$ is the loss function for $m$-th tree, $T_m$, defined in Eq.~\eqref{eq_obj}. 

In the testing phase, the output of a forest is the average of outputs of all the decision trees in the forest, which can be defined as follows:
\begin{equation}
\vct{g}(\vct{x}; \vct\theta, \mathcal{F}) = \frac{1}{M}\sum_{m=1}^{M}\vct{g}(\vct{x};\vct\theta, T_m).
\end{equation}

\subsection{Optimization}\label{learning_split_node}

Since a forest is a collection of decision trees, this subsection presents how to optimize a decision tree, which can then be extended to optimize a forest. 

It is obvious that the construction of an ordinal regression tree includes two parts of parameters---the parameter $\vct{\theta}$ in convolutional neural networks for the split nodes and the parameter $\vct{\tau}$ for ordinal distribution in the leaf nodes. The optimal parameters $(\vct{\theta}^\star, \vct{\tau}^\star)$ are obtained by minimizing the objective function in Eq.~\eqref{eq_simple_obj}: 
\begin{equation}
(\vct{\theta}^\star, \vct{\tau}^\star) = \argmin_{\vct{\theta}, \vct{\tau}}\ell_{\mathcal{S}}(\vct{\theta}, \vct{\tau}).
\label{best_obj}
\end{equation}
A common way to solve Eq.~\eqref{best_obj} is to optimize $\vct{\theta}$ and 
$\vct{\tau}$  alternatively. Therefore, the following presents the derivatives of the objective function $\ell_{\mathcal{S}}$ with respect to $\vct{\theta}$ and $\vct{\tau}$, and how to extend it to a forest.

\subsubsection{Optimizing \texorpdfstring{$\vct{\theta}$}{}}

Here, we present how to derive the derivative of objective function $\ell_{\mathcal{S}}$ with respect to the parameter $\vct{\theta}$ for split nodes, when the ordinal distribution $\vct{\tau}$ in leaf nodes is fixed. The derivative of $\ell_{\mathcal{S}}(\vct{\theta}, \vct{\tau})$ with respect to $\vct{\theta}$ can be calculated as follows:
\begin{align}\label{split_node_function}
\frac{\partial\ell_{\mathcal{S}}(\vct{\theta}, \vct{\tau})}{\partial\vct{\theta}} 
= \sum_{i=1}^{N}\sum_{n\in\mathcal{N}}\frac{\partial\ell_{\mathcal{S}}(\vct{\theta}, \vct{\tau})}{\partial \vct{f}_{\vct{\theta}}(\vct{x}_i)_{\gamma{(n)}}}\frac{\partial \vct{f}_{\vct{\theta}}(\vct{x}_i)_{\gamma{(n)}}}{\partial\vct{\theta}},
\end{align}
where the first  and second term on the right-hand side rely on the tree structure $T$ and the index function $\gamma(n)$ of $\vct{f}_{\vct{\theta}}$, respectively. 

The first term in Eq.~\eqref{split_node_function} can be obtained as follows:
\begin{align}\label{first_part}
&\frac{\partial \ell_{\mathcal{S}}(\vct{\theta}, \vct{\tau})}{\partial \vct{f}_{\vct\theta}(\vct{x}_i)_{\gamma{(n)}}}  \\
=& \sum_{l\in\mathcal{L}}\frac{\partial \ell_{\mathcal{S}}(\vct{\theta}, \vct{\tau})}{\partial p(l|\vct{x}_i;\vct\theta)}\frac{\partial p(l|\vct{x}_i;\vct\theta)}{\partial \vct{f}_{\vct\theta}(\vct{x}_i)_{\gamma{(n)}}} \notag \\
=& -\frac{1}{N}\sum_{k=1}^{K-1}\sum_{c=1}^{2}\sum_{l\in\mathcal{L}}
\frac{d_i^{(k,c)}\tau_{l}^{(k,c)}}{\sum_{l\in\mathcal{L}}\left(p(l|\vct{x}_i;\vct\theta)\tau_{l}^{(k,c)}\right)} \frac{\partial p(l|\vct{x}_i;\vct\theta)}{\partial \vct{f}_{\vct\theta}(\vct{x}_i)_{\gamma{(n)}}} \notag \\
=& -\frac{1}{N}\sum_{k=1}^{K-1}\sum_{c=1}^{2}\sum_{l\in\mathcal{L}}
\frac{d_i^{(k,c)}\tau_{l}^{(k,c)}p(l|\vct{x}_i;\vct\theta)}{\sum_{l\in\mathcal{L}}\left(p(l|\vct{x}_i;\vct\theta)\tau_{l}^{(k,c)}\right)} \frac{\partial \log p(l|\vct{x}_i;\vct\theta)}{\partial \vct{f}_{\vct\theta}(\vct{x}_i)_{\gamma{(n)}}}.\notag
\end{align}
With the definitions of $p(l|\vct{x};\vct\theta)$ and $s_n(\vct{f}_{\vct{\theta}}(\vct{x}))$ in Sec.~\ref{sec_decision_tree},  we have
\begin{align}\label{later_part}
&\frac{\partial \log p(l|\vct{x}_i;\vct\theta)}{\partial \vct{f}_{\vct\theta}(\vct{x}_i)_{\gamma{(n)}}} \notag \\
=& \frac{\partial\log \left(s_n(\vct{f}_{\vct{\theta}}(\vct{x}_i))^{\vct{1}[l\in\mathcal{L}_n^l]}\right)}{\partial \vct{f}_{\vct\theta}(\vct{x}_i)_{\gamma{(n)}}} + 
\frac{\partial\log\left(1 - s_n\left(\vct{f}_{\vct{\theta}}(\vct{x}_i)\right)^{\vct{1}[l\in\mathcal{L}_n^r]}\right)}{\partial \vct{f}_{\vct\theta}(\vct{x}_i)_{\gamma{(n)}}} \notag \\
=& \left(1 - s_n(\vct{f}_{\vct{\theta}}(\vct{x}_i))^{\vct{1}[l\in\mathcal{L}_n^l]}\right) - s_n(\vct{f}_{\vct\theta}(\vct{x}_i))^{\vct{1}[l\in\mathcal{L}_n^r]}.
\end{align}
Substituting Eq.~\eqref{later_part} into Eq.~\eqref{first_part} yields
\begin{align}\label{eq_update_theta}
&\frac{\partial \ell_{\mathcal{S}}(\vct{\theta}, \vct{\tau})}{\partial \vct{f}_{\vct\theta}(\vct{x}_i)_{\gamma{(n)}} } \notag \\
=& \frac{1}{N}\sum_{k=1}^{K-1}\sum_{c=1}^{2}d_i^{(k,c)}\Big(
\frac{\sum_{l\in\mathcal{L}}p(l|\vct{x}_i;\vct\theta)\tau_{l}^{(k,c)}s_n(\vct{f}_{\vct\theta}(\vct{x}_i))^{\vct{1}[l\in\mathcal{L}_n^r]}}{\sum_{l\in\mathcal{L}}\left(p(l|\vct{x}_i;\vct\theta)\tau_{l}^{(k,c)}\right)}  \notag \\
&\quad\quad\quad-\frac{\sum_{l\in\mathcal{L}}p(l|\vct{x}_i;\vct\theta)\tau_{l}^{(k,c)}\left(1 - s_n(\vct{f}_{\vct\theta}(\vct{x}_i))^{\vct{1}[l\in\mathcal{L}_n^l]}\right)}{\sum_{l\in\mathcal{L}}\left(p(l|\vct{x}_i;\vct\theta)\tau_{l}^{(k,c)}\right)}\Big) \notag \\
=& \frac{1}{N}\sum_{k=1}^{K-1}\sum_{c=1}^{2}d_i^{(k,c)}\Big(s_n(\vct{f}_{\vct\theta}(\vct{x}_i))
\frac{\sum_{l\in\mathcal{L}^r}p(l|\vct{x}_i;\vct\theta)\tau_{l}^{(k,c)}}{\sum_{l\in\mathcal{L}}\left(p(l|\vct{x}_i;\vct\theta)\tau_{l}^{(k,c)}\right)}  \notag \\
&\quad\quad\quad\quad-(1 - s_n(\vct{f}_{\vct\theta}(\vct{x}_i)))\frac{\sum_{l\in\mathcal{L}^l}p(l|\vct{x}_i;\vct\theta)\tau_{l}^{(k,c)}}{\sum_{l\in\mathcal{L}}\left(p(l|\vct{x}_i;\vct\theta)\tau_{l}^{(k,c)}\right)}\Big) \notag \\
&= \frac{1}{N}\sum_{k=1}^{K-1}\sum_{c=1}^{2}d_i^{(k,c)}\Big(s_n(\vct{f}_{\vct\theta}(\vct{x}_i))
\frac{g_k^c(\vct{x}_i;\vct\theta, T^r)}{g_k^c(\vct{x}_i;\vct\theta,T)}  \notag \\
&\quad\quad\quad\quad\quad\quad-(1 - s_n(\vct{f}_{\vct\theta}(\vct{x}_i)))\frac{g_k^c(\vct{x}_i;\vct\theta, T^l)}{g_k^c(\vct{x}_i;\vct\theta, T)}\Big),
\end{align}
where 
\begin{align}
\begin{cases}
g_k^c(\vct{x}_i;\vct\theta,T^l) = \sum\limits_{l\in\mathcal{L}^l}p(l|\vct{x}_i;\vct\theta)\tau_{l}^{(k,c)},\\
g_k^c(\vct{x}_i;\vct\theta,T) = \sum\limits_{l\in\mathcal{L}}p(l|\vct{x}_i;\vct\theta)\tau_{l}^{(k,c)}.
\end{cases}
\end{align}
We note that $g_k^c(\vct{x}_i;\vct\theta,T) = g_k^c(\vct{x}_i;\vct\theta,T^l) + g_k^c(\vct{x}_i;\vct\theta,T^r)$, implying that the derivative in Eq.~\eqref{eq_update_theta} can be computed from the leaf nodes to the root in a bottom-up manner. Thus, the parameter $\vct\theta$ used in split node can be updated through standard back-propagation.

\subsubsection{Optimizing \texorpdfstring{$\vct{\tau}$}{}}
After updating the parameter $\vct{\theta}$ through Eq.~\eqref{eq_update_theta}, we turn to optimize $\vct{\tau}$. With $\vct{\theta}$ being fixed, the overall objective function in Eq.~\eqref{eq_simple_obj} reduces to
\begin{equation}
\min_{\vct{\tau}} \ell_{\mathcal{S}}(\vct{\theta}, \vct{\tau}), \quad\mathrm{s.t.}\quad\forall \ l,k,\  \sum_{c=1}^{2}\tau_{l}^{(k,c)} = 1,
\end{equation}
which is a constrained optimization problem. Following~\cite{jordan1999introduction,yuille2003concave}, we resort to \emph{variational bounding} to solve the above optimization problem. Similar to~\cite{shen2017label}, an upper bound for the loss function $\ell_{\mathcal{S}}(\vct{\theta}, \vct{\tau})$, with $\vct{\theta}$ being fixed, can be derived by Jensen's inequality:
\begin{align}\label{jensen_inequality_eq1}
&\ell_{\mathcal{S}}(\vct{\theta}, \vct{\tau}) \\
=& -\frac{1}{N}\sum_{i=1}^{N}\sum_{k=1}^{K-1}\sum_{c=1}^{2}d_i^{(k,c)}\log\Big(\sum_{l\in\mathcal{L}}p(l|\vct{x}_i;\vct{\theta})\tau_{l}^{(k,c)}\Big) \notag \\
\le& -\frac{1}{N}\sum_{i=1}^{N}\sum_{k=1}^{K-1}\sum_{c=1}^{2}d_i^{(k,c)}\sum_{l\in\mathcal{L}}\xi_l(\bar\tau_{l}^{(k,c)}, \vct{x}_i)\notag\\
&\quad\quad\quad\quad\quad\quad\quad\quad\quad\quad\quad\quad\times\log\Big(\frac{p(l|\vct{x}_i;\vct\theta)\tau_{l}^{(k,c)}}{\xi_l(\bar\tau_l^{(k,c)}, \vct{x}_i)}\Big),\notag
\end{align}
where $\xi_l(\tau_l^{(k,c)}, \vct{x}_i) = \tfrac{p(l|\vct{x}_i;\vct{\theta})\tau_l^{(k,c)}}{\sum_{l\in\mathcal{L}}p(l|\vct{x}_i;\vct{\theta})\tau_l^{(k,c)}}$. 

By defining the right-hand side of the inequation in Eq.~\eqref{jensen_inequality_eq1} as $\phi(\vct{\bar\tau}, \vct{\tau})$,
we know that $\phi(\vct{\bar\tau}, \vct{\tau})$ is an upper bound for the objective function $\ell_{\mathcal{S}}(\vct{\theta}, \vct{\tau})$.
This implies that for any $\vct{\tau}$ and $\vct{\bar\tau}$, we have $\ell_{\mathcal{S}}(\vct{\theta}, \vct{\tau}) \leq \phi(\vct{\bar\tau}, \vct{\tau})$, and this equality holds if and only if $\vct{\bar\tau} = \vct{\tau}$. Here we use $\vct{\tau}^{(t)}$ and $\vct{\tau}^{(t+1)}$ to denote the ordinal distribution at the $t$- and $(t+1)$-th iterations, respectively. It is obvious that $\phi(\vct{\tau}^{(t)}, \vct{\tau})$ is an upper bound for the objective function $\ell_{\mathcal{S}}(\vct{\theta}, \vct{\tau})$. Moreover, there is at least one point $\vct{\tau}^{(t+1)}$ satisfying the inequality
$\phi( \vct{\tau}, \vct{\tau}^{(t+1)}) \le \ell_{\mathcal{S}}(\vct{\theta}, \vct{\tau}^{(t)})$. This further implies $\ell_{\mathcal{S}}(\vct{\theta}, \vct{\tau}^{(t+1)}) \leq \ell_{\mathcal{S}}(\vct{\theta}, \vct{\tau}^{(t)})$, because $ \phi( \vct{\tau}, \vct{\tau}^{(t+1)}) = \ell_{\mathcal{S}}(\vct{\theta}, \vct{\tau}^{(t+1)})$. Therefore, we can replace $\ell_{\mathcal{S}}(\vct{\theta}, \vct{\tau})$ with $\phi(\vct{\bar{\tau}}, \vct{\tau})$ as the objective function after guaranteeing that $\ell_{\mathcal{S}}(\vct{\theta}, \vct{\tau}^{(t)})=\phi(\vct{\bar{\tau}}, \vct{\tau}^{(t)})$; \ie, $\vct{\bar\tau} = \vct{\tau}^{(t)}$. Then we have  
\begin{align}
\vct{\tau}^{(t+1)} = \argmin_{\vct{\tau}}\phi(\vct{\tau}^{(t)}, \vct{\tau})\ \mathrm{s.t.}\ \forall\ l,k, \sum_{c=1}^{2} \tau_l^{(k,c)}= 1.
\end{align}
This can be solved by minimizing the Lagrangian defined as
\begin{equation}
\psi(\vct{\tau}^{(t)}, \vct{\tau}) = \phi(\vct{\tau}^{(t)}, \vct{\tau}) \!+\! \sum_{k=1}^{K-1}\sum_{l\in\mathcal{L}}\lambda_{l}^{k}\left(\sum_{c=1}^{2}\tau_l^{(k,c)} \!-\! 1\right),
\end{equation}
where $\lambda_{l}^{k}$ represents the Lagrange multiplier. 
By letting $\partial\psi(\vct{\tau}^{(t)}, \vct{\tau})/\partial\tau_l^{(k,c)} = 0$, we have
\begin{align}
\lambda_{l}^{k} &= \frac{1}{N}\sum_{i=1}^{N}\sum_{c=1}^{2}d_i^{(k,c)}\xi_l({\tau_{l}^{(k,c)}}^{(t)}, \vct{x}_i), \notag\\
{\tau_{l}^{(k,c)}}^{(t+1)} &= \frac{\sum_{i=1}^{N}d_i^{(k,c)}\xi_l({\tau_{l}^{(k,c)}}^{(t)}, \vct{x}_i)}{\sum_{c=1}^{2}\sum_{i=1}^{N}d_i^{(k,c)}\xi_l({\tau_{l}^{(k,c)}}^{(t)}, \vct{x}_i)}.
\end{align}
Then the ${\tau_{l}^{k}}^{(t+1)}$ can be updated by:
\begin{align}
&{\tau_{l}^{k}}^{(t+1)} = {\tau_{l}^{(k,c=1)}}^{(t+1)} \notag\\
&= \frac{\sum\limits_{i=1}^{N}d_i^{(k,c=1)}\xi_l({\tau_{l}^{(k,c=1)}}^{(t)}, \vct{x}_i)}{\sum\limits_{i=1}^{N}\!d_i^{(k,c=1)}\!\xi_l({\tau_{l}^{(k,c=1)}}^{(t)},\! \vct{x}_i)\! +\! \sum\limits_{i=1}^{N}\!d_i^{(k,c=2)}\!\xi_l({\tau_{l}^{(k,c=2)}}^{(t)},\! \vct{x}_i)} \notag \\
&= \frac{\sum\limits_{i=1}^{N}d_i^k\xi_l({\tau_{l}^{k}}^{(t)}, \vct{x}_i)} {\sum\limits_{i=1}^{N}d_i^k\xi_l({\tau_{l}^{k}}^{(t)}, \vct{x}_i) \!+\! \sum\limits_{i=1}^{N}(1 \!-\! d_i^k)\xi_l((1 - {\tau_{l}^{k}}^{(t)}), \vct{x}_i)}.
\label{eq_update_leaf_node}
\end{align}
Eq.~\eqref{eq_update_leaf_node} can be used to update the ordinal distribution in the leaf nodes. The ordinal distribution at the starting point $\vct{\tau}_l^{(0)}$ can be arbitrary as long as all elements are positive and within $[0,1]$. In this paper, we initialize ${\tau_{l}^{k}}^{(0)}$ as $1/2$, for $\forall\ l\in\mathcal{L}, k\in \{1,2,\ldots,K-1\}$.

\begin{algorithm}[t]
\small
\caption{The pseudo code for training our CORF.}
\label{alg_CORF}
\begin{algorithmic}[1]
\STATE{\textbf{Input}: $\mathcal{S}$: training set;\\
\quad\quad\ $n_{\vct\theta}$: no. of mini-batch to update $\vct\theta$;\\
\quad\quad\ \  $n_{\vct\tau}$: no. of iterations to update $\vct\tau$}.
\STATE{\textbf{Initialization}: Initialize $\vct\theta$ randomly, let ${\tau_{l}^{k}}^{(0)} = \frac{1}{2}$ for $\forall\ l\in\mathcal{L}, k\in \{1,2,\ldots,K-1\}$, and set $\bar{\mathcal{B}}=\{\emptyset\}$.}
\WHILE{Not converge}
\WHILE{$|\bar{\mathcal{B}}|< n_{\vct\theta}$}
\STATE{Randomly select a mini-batch $\mathcal{B}$ from $\mathcal{S}$}
\STATE{Compute gradient (Eq.~\eqref{update_theta}) on $\mathcal{B}$ to update $\vct\theta$}
\STATE{$\bar{\mathcal{B}} = \bar{\mathcal{B}}\cup \mathcal{B}$}
\ENDWHILE
\STATE{Iterate Eq.~\eqref{eq_update_leaf_node} on $\bar{\mathcal{B}}$ to update $\vct\tau$, and repeat $n_{\vct{\tau}}$ times}
\STATE{$\bar{\mathcal{B}} = \{\emptyset\}$}
\ENDWHILE
\end{algorithmic}
\end{algorithm}

\subsubsection{Extension to the Forest}

Extending the optimization for ordinal regression tree to a forest is straightforward. After fixing the ordinal distribution $\vct{\tau}$ in the leaf nodes of each tree in the forest $\mathcal{F}$, the derivative of the overall loss function with respect to $\vct{\theta}$ can be obtained as:
\begin{equation}
\label{update_theta}
\frac{\partial\ell^\mathcal{F}}{\partial\vct{\theta}} = \frac{1}{M}\!\sum_{i=1}^{N}\!\sum_{m=1}^{M}\!\sum_{n\in\mathcal{N}_m}\!\frac{\partial \ell^{{T}_m}}{\partial \vct{f}_{\vct{\theta}}(\vct{x}_i)_{\gamma_{m(n)}}} \frac{\partial \vct{f}_{\vct{\theta}}(\vct{x}_i)_{\gamma_{m(n)}}}{\partial\vct{\theta}},
\end{equation}
where $\gamma_m(\cdot)$ is the index function and  $\mathcal{N}_m$ is set of the split nodes of the $m$-th tree, $T_m$. Since the index function $\gamma_m(\cdot)$ of each tree is randomly initialized before the training process, as a result, the split nodes of a tree correspond to a subset of output units of $\vct{f}_{\vct\theta}(\cdot)$. This strategy can reduce the risk of overfitting by introducing the randomness in training.

When $\vct{\theta}$ is fixed, we can update the ordinal distribution in each leaf node independently since all trees are independent. For implementation convenience, the ordinal distributions $\vct{\tau}$ in  leaf nodes are updated based on a set of mini-batches of dataset, $\bar{\mathcal{B}}$, instead of the whole dataset, $\mathcal{S}$. Algorithm~\ref{alg_CORF} presents the training procedure of our CORF.

\section{Experiments}\label{exp}
To demonstrate the effectiveness of CORF, we verify it on two image ordinal estimation tasks, \emph{i.e.}, facial age estimation and image aesthetic assessment.

\subsection{Datasets}

\subsubsection{Facial Age Estimation Datasets}
FGNET~\cite{panis2016overview} is a database used for age estimation, which consists of only 1,002 color or gray facial images of 82 subjects with large variations in pose, expression, and lighting. For each subject, there are more than ten images ranging from age 0 to age 69. MORPH~\cite{ricanek2006morph} is one of the most popular facial age estimation dataset, which contains 55,134 facial images of 13,617 subjects ranging from 16 to 77 years old. CACD~\cite{chen2015face} is another largest facial age estimation database, which contains around 160,000 facial images of 2,000 celebrities. All facial images are aligned by the five facial landmarks detected by MTCNN~\cite{zhang2016joint}. Following~\cite{pan2018mean}, we use the leave-one-person-out (LOPO) protocol in the FGNET dataset and employ the five-fold random split (RS) protocol and the five-fold subject-exclusive (SE) protocol in MORPH dataset. According to~\cite{shen2018deep}, CACD is divided into three subsets: the training set composed of 1,800 celebrities, the testing set that has 120 celebrities and the validation set containing 80 celebrities. The partitioning interval $\eta$ is set as $1$ in three facial age estimation databases.

\begin{figure}[t]
\setlength\abovecaptionskip{-0.7\baselineskip}
\centering
\includegraphics[width=1\linewidth]{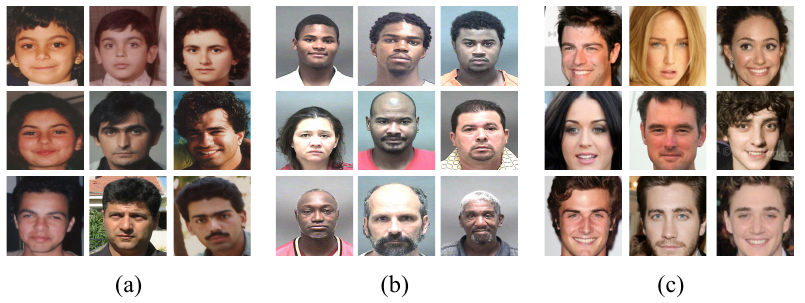}
\caption{Some facial image samples from three age estimation databases, including (a) FGNET, (b) MORPH, and (c) CACD.}
\label{facial_sample}
\end{figure}

\subsubsection{Image Aesthetic Assessment Datasets}
AVA~\cite{murray2012ava} is the largest image aesthetics assessment dataset and contains more than 250,000 pictures that are downloaded from well-known photographer community sites named DPChallenge. The aesthetic quality of each image is rated by a score vector which ranges from 1 to 10. In our task, we average the score vector and convert the average score into an ordinal distribution with a partitioning interval of $0.1$. Followed by~\cite{murray2012ava}, we adopt the same setting to split the dataset.  AADB~\cite{kong2016photo} is another common image aesthetics assessment dataset. Different from the AVA dataset, AADB contains 10,000 pictures labeled by scalars representing their average aesthetic score collected from Flickr with the score ranges from 0 to 1 and all pictures are the natural images without artificial modification. In this work, the 9,000 of 10,000 pictures are used for training and the rest are for testing. The partitioning interval is set as $0.05$ in the AADB dataset.
\begin{figure}[t]
\setlength\abovecaptionskip{-0.7\baselineskip}
\centering
\includegraphics[width=1\linewidth]{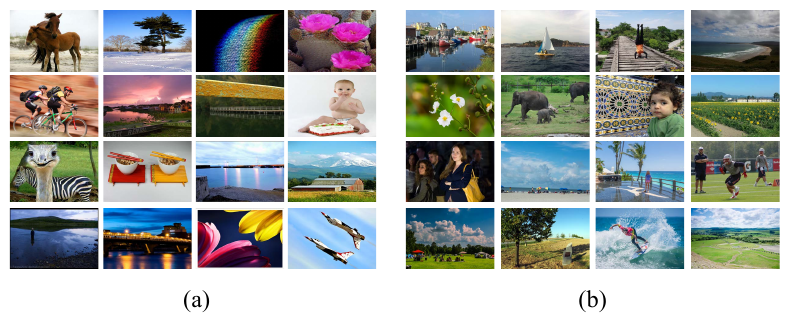}
\caption{Some aesthetic assessment samples from the (a) AVA dataset and (b)  AADB dataset.}
\label{aesthetic_sample}
\end{figure}
Before fed into the networks, all the input images are randomly cropped into $256 \times 256 \times 3$ and then resized into $224 \times 224 \times 3$. In order to better understand these dataset, we show some facial age estimation samples and some aesthetic assessment samples in Figs.~\ref{facial_sample} and~\ref{aesthetic_sample}, respectively.

\begin{table*}[h]
\renewcommand\arraystretch{1.2}
\caption{Performance comparisons between the CORF and the state-of-the-arts on the FGNET, MORPH, and CACD datasets. Note that the bold and underlined values represent the best and the second best performance, respectively.
}\label{age_estimation_result}
\centering
\begin{tabular}{cccccccc}
\toprule
Dataset  & Evaluation  & OHRank\cite{chang2011ordinal} & OR-CNN\cite{niu2016ordinal}  & Ranking-CNN\cite{chen2017using} & Mean-Variance\cite{pan2018mean} & DRFs\cite{shen2018deep}      & CORF (Ours)   \\ \midrule
\multirow{2}{*}{FGNET}  & MAE & 4.48   & N/A & N/A  & 4.10  & \underline{3.85}  & \bf{2.68}     \\ 
           & CS($L=5$) & 74.4\% & N/A & N/A  & 78.2\%  & \underline{80.6\%} & \bf{86.80\%}  \\ \hline
\multirow{2}{*}{MORPH}& MAE  & 3.82   & 3.27    & 2.96  & 2.41  & \bf{2.17} & \underline{2.19} \\ 
        & CS($L=5$) & N/A    & 73.0\% & 85.0\%     & 90.0\%   & \underline{91.3\%}    & \bf{93.0\%}   \\ \hline   
\multirow{2}{*}{CACD}    & MAE (train)  & N/A  & 4.89  & N/A  & \bf{4.64}  & \bf{4.64} & \underline{4.67}          \\ 
                         & MAE (val)    & N/A    & 6.87    & N/A         & 6.29          & \bf{5.77} & \underline{6.10}          \\ \bottomrule
\end{tabular}
\end{table*}

\subsection{Implementation Details}
Since the CORF is modular and its feature function can be implemented as a standard neural network layer, we can integrate it with any deep networks. Following the recent CNN-based methods~\cite{shen2018deep,rothe2018deep,pan2018mean,tian2018training}, we use the VGG-16 that is pre-trained with ImageNet~\cite{russakovsky2015imagenet} as the backbone of our method. The default parameter settings for our method are: 
number of trees (5), tree depth (6), number of output units in the feature learning function (128), number of iterations to update leaf node predictions (20), number of mini-batches used to update leaf node predictions (10). The network training based hyper-parameters are: initial learning rate (0.001), mini-batch size (64), maximal epochs (40), and the probability of dropout is 0.5. We decrease the learning rate ($\times 0.2$) every 15 epochs. 
To further validate the effectiveness of the CORF, we implement a convolutional ordinal regression (COR) method as a baseline. COR utilizes VGG-16 as its backbone and the output layer is the same as the OR-CNN~\cite{niu2016ordinal}.

\subsection{Evaluation on Facial Age Estimation}
We evaluate the CORF on three public facial age estimation datasets, \emph{i.e.}, FGNET, MORPH, and CACD, and compare it with several ordinal regression methods (\eg, OHRank\cite{chang2011ordinal}, OR-CNN\cite{niu2016ordinal}, and Ranking-CNN\cite{chen2017using}) and the state-of-the-art methods of age estimation (\eg, Mean-Variance\cite{pan2018mean} and DRFs\cite{shen2018deep}). Further, we evaluate the performance of age estimation by two measurements, \ie, Mean Absolute Error (MAE) and the Cumulative Score (CS).

\begin{table*}[t]
\renewcommand\arraystretch{1.2}
\caption{Comparisons between COR and CORF in terms of MAE and CS.}\label{OR_VS_ORFs}
\centering
\begin{tabular}{ccccccccc}
    \toprule
    \multirow{3}{*}{Method} & \multicolumn{2}{c}{FGNET} & & \multicolumn{5}{c}{MORPH}\\ \cline{2-3}\cline{5-9}
                            & \multicolumn{2}{c}{Leave-One-Person-out (LOPO)} & & \multicolumn{2}{c}{Random Split (RS)} & & \multicolumn{2}{c}{Subject Exclusive (SE)}\\   \cline{2-3}\cline{5-6}\cline{8-9}
                            & MAE   & CS ($L=5$) && MAE  & CS ($L=5$) && MAE   & CS ($L=5$) \\  \midrule
    COR                     &  3.49 & 80.87\%    && 2.30 & 91.60\%     && 2.77  & 87.50\% \\ \hline 
    CORF (Ours)            &  \bf{2.68} & \bf{86.80\%}    && \bf{2.19} & \bf{93.00\%}     && \bf{2.69}  & \bf{89.00\%} \\ \bottomrule
\end{tabular}
\end{table*}

\begin{table*}[t]
\renewcommand\arraystretch{1.2}
\caption{Performance comparisons between the proposed CORF and the state-of-the-art methods on the AADB and the AVA datasets. Note that the bold and underlined values represent the best and the second best performance, respectively.}\label{aestab}
\centering
\begin{tabular}{ccccccccc}

\toprule
Dataset & Evaluation & Reg+Rank\cite{kong2016photo} & EMD\cite{hou2016squared} & NIMA-V1\cite{talebi2018nima} & GPF-CNN\cite{zhang2019gated} & OR-CNN\cite{niu2016ordinal} & COR & CORF (Ours) \\ \midrule
 \multirow{2}*{AADB} &PLCC & N/A & N/A & N/A & N/A & 0.4388 & \underline{0.6555} &  $\mathbf{0.6829}$  \\
                     &SRCC & 0.6308 & \underline{0.6647} & N/A & N/A & 0.437 & 0.6416 & $\mathbf{0.6770}$ \\ \hline
 \multirow{2}*{AVA}  &PLCC & N/A & N/A & 0.610 & $\mathbf{0.6868}$ & 0.5076 & 0.6218 & \underline{0.6649} \\
                &SRCC & 0.5126 & N/A & 0.592 & $\mathbf{0.6762}$ & 0.4986 & 0.6115 & \underline{0.6714} \\ \bottomrule
\end{tabular}
\end{table*}

The results of CORF and the competitors are summarized in Table~\ref{age_estimation_result}. It can be seen that CORF outperforms three state-of-the-art ordinal regression methods (OHRank, OR-CNN, and Ranking-CNN) and achieves a comparable performance with two state-of-the-art age estimation methods (Mean-Variance and DRFs) on the facial age estimation datasets. This result can verify the effectiveness and superiority of the proposed method over the existing ordinal regression methods to a certain extent.
We note that the performance of our CORF is close to the one of DRFs and both of them achieve the best performance on facial age estimation task. The reasons are twofold. First, both of them are based on the differentiable decision trees, which is an integrated method. Second, they both can well learn the inner relationships of the age patterns, even if their underlying principles and assumptions are different. However, the DRFs are a label distribution learning approach based on the hypothesis of Gaussian distribution, while our CORF is an ordinal regression approach based on the hypothesis of ordinal distribution. Both of them can achieve a promising performance when the data are sufficient, such as on the MORPH dataset and CACD dataset. Actually, the results of our CORF are not exactly the same as those of the DRFs. Our approach is far superior to the performance of DRFs on the FGNET dataset, which is the smallest dataset. The reason is that it is easier to learn an ordinal relationship between ages directly than to learn a Gaussian distribution in a small dataset.

\begin{figure}[t]
\setlength\abovecaptionskip{-0.7\baselineskip}
\centering
\includegraphics[width=1\linewidth]{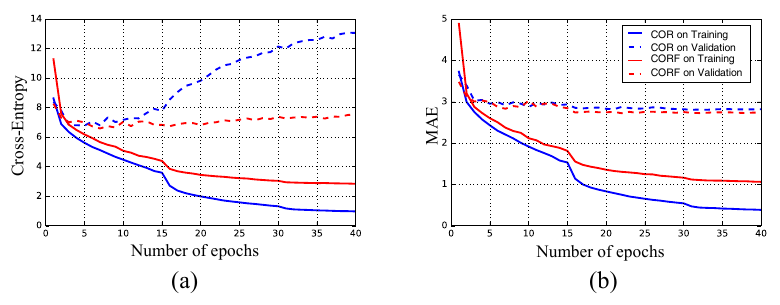}
\caption{The train and validation losses on MORPH (SE), \ie, (a) cross-entropy loss and (b) MAE loss.}\label{analysis_vis_loss}
\end{figure}

To further verify the effectiveness of CORF, we compare it with COR on FGNET and MORPH datasets. In order to eliminate the impact of network parameters and network size on performance, both COR and CORF utilize VGG-16 as their backbone. From Table~\ref{OR_VS_ORFs}, we learn that CORF performs better than COR in all situations. It demonstrates that learning a global ordinal distribution can obtain better predictive performance than learning a series of binary classifiers independently.

Moreover, Fig.~\ref{analysis_vis_loss} (a) shows that COR is more prone to overfitting than CORF on cross-entropy loss, which means CORF is more stable. Fig.~\ref{analysis_vis_loss} (b) shows that the over-fitting issue of COR is less obvious in MAE loss, because the calculation manner of MAE in Eq.~\eqref{cal_final_output} may have some fault tolerance. In conclusion, CORF is more effective and stable than COR, because learning an ordinal distribution can better preserve the global ordinal relationship among these binary classifiers and combining with an ensemble method, \ie, decision tree, make our model more stable.

\subsection{Evaluation on Image Aesthetic Assessment}
We validate our CORF on two image aesthetics assessment datasets, \ie, AVA and AADB, by comparing with the state-of-the-art aesthetics assessment methods and several ordinal regression methods. Further, Pearson Linear Correlation Coefficient (PLCC) and Spearman’s Rank Correlation Coefficient (SRCC) are computed between the predicted and ground truth aesthetics mean scores for measuring the correlation-based performance of image aesthetics assessment. We know that the PLCC is a statistic that measures linear correlation between two variables and it has a value between 1 and $-1$, where 1 is total positive linear correlation, 0 is no linear correlation, and $-1$ is total negative linear correlation. Similar to the PLCC, the SRCC can be defined as the Pearson correlation coefficient between two rank variables.

From the results of Table~\ref{aestab}, it can be seen that CORF outperforms the current image aesthetics assessment methods and other ordinal regression methods on AADB dataset. Besides, our method achieves a comparable performance on AVA dataset. The GPF-CNN outperforms our method, but it is more complicated both in data processing and backbone network. To sum up, the proposed method is effective on image aesthetic assessment.

\begin{figure}[t]
\setlength\abovecaptionskip{-0.7\baselineskip}
\centering
\includegraphics[width=1\linewidth]{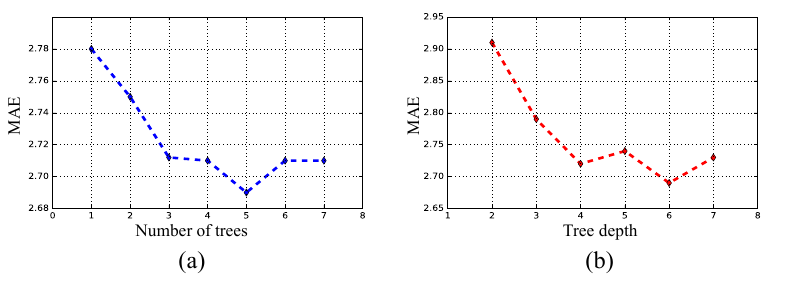}
\caption{The performance changes by varying (a) number of tree and (b) tree depth on MORPH (SE).}
\label{analysis_vis_tree}
\end{figure}

\begin{figure}[t]
\setlength\abovecaptionskip{-0.7\baselineskip}
\centering
\includegraphics[width=1\linewidth]{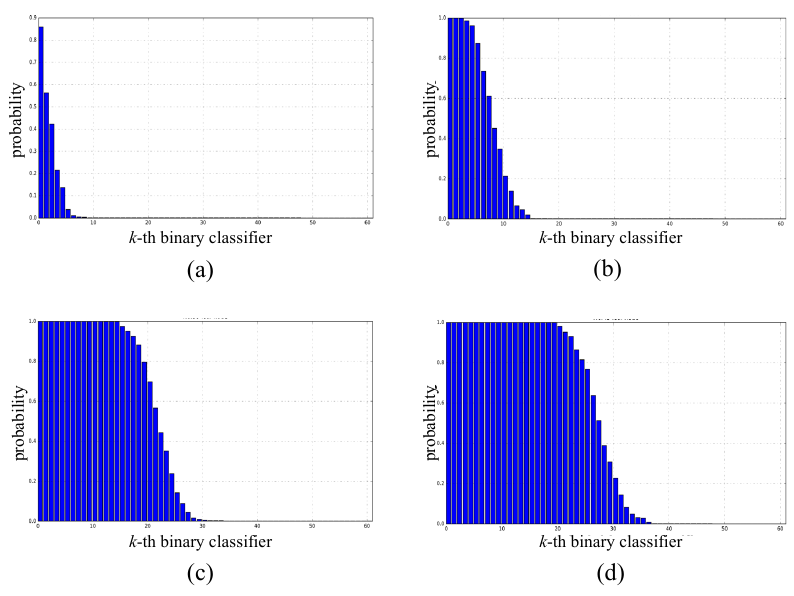}
\caption{The visualization of leaf nodes. These nodes from (a) to (d) represent the age distributions of juvenile, youth, middle-aged, and old age, respectively.}
\label{leaf_node}
\end{figure}
\begin{figure}[t]
\setlength\abovecaptionskip{-0.7\baselineskip}
\centering
\includegraphics[width=1\linewidth]{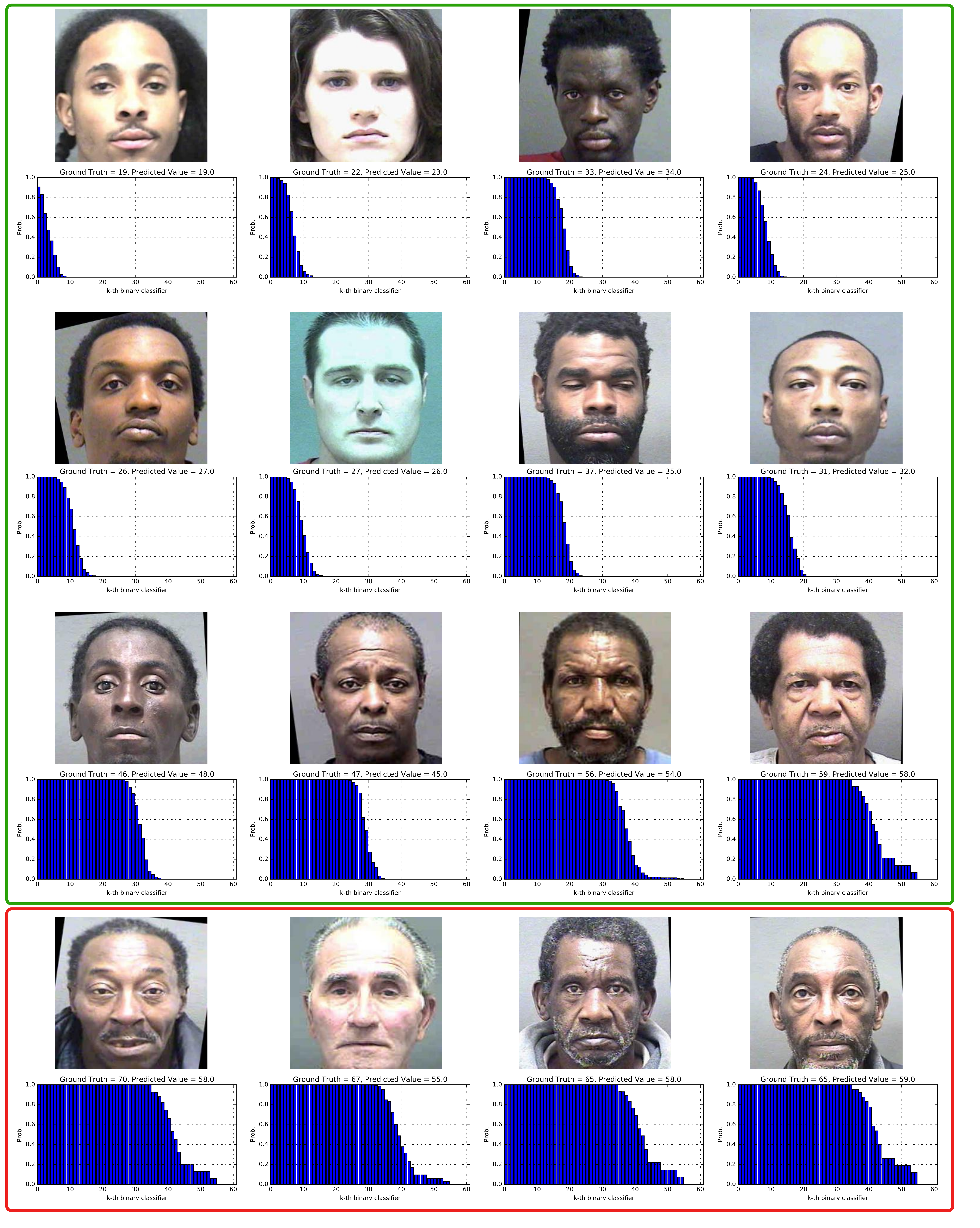}
\caption{The age label range of MORPH is $[16, 77]$ and each age label is transformed into a 61 dimensionality ordinal distribution label with a partitioning interval 1.}
\label{age_estimation_vis}
\end{figure}

\subsection{Effects on the Number of Trees and Tree Depth}
As an ensemble model, the proposed CORF has two important hyper-parameters: number of tree and tree depth. Now we change each of them and fix the other one to the default value (number of tree (5) and tree depth (6)) to see how the performance changes on MORPH (SE). As shown in Fig.~\ref{analysis_vis_tree} (a), the more trees we have, the better performance the CORF will have. Besides, with the tree depth increase, as shown in Fig.~\ref{analysis_vis_tree} (b), the MAE first starts to decrease and then becomes stable.

\subsection{Visualization of the Leaf Nodes}
To better understand the CORF, we visualize the output of leaf nodes learned on MORPH. We can see that different leaf node learns different ordinal distribution, \eg, leaf node from Fig.~\ref{leaf_node} (a) to (d) learns an age distribution of juvenile, youth, middle-aged, and old age, respectively. The distribution diversity of leaf nodes is necessary to model any desired ordinal distribution by a mixture of leaf nodes. According to the update rule of the leaf node  in Eq.~\eqref{eq_update_leaf_node}, the update of leaf node distribution is a weighted combination of the training sample labels. Since the labels of the training sample are all in descending order, \eg, the label 5 years old can be transformed into an ordinal distribution label $[1,1,1,1,0,...,0]\T$ with descending order. Thus the distribution of the leaf node can also preserve the descending order after the weighted combination. 

\begin{figure}[t]
\setlength\abovecaptionskip{-0.7\baselineskip}
\centering
\includegraphics[width=1\linewidth]{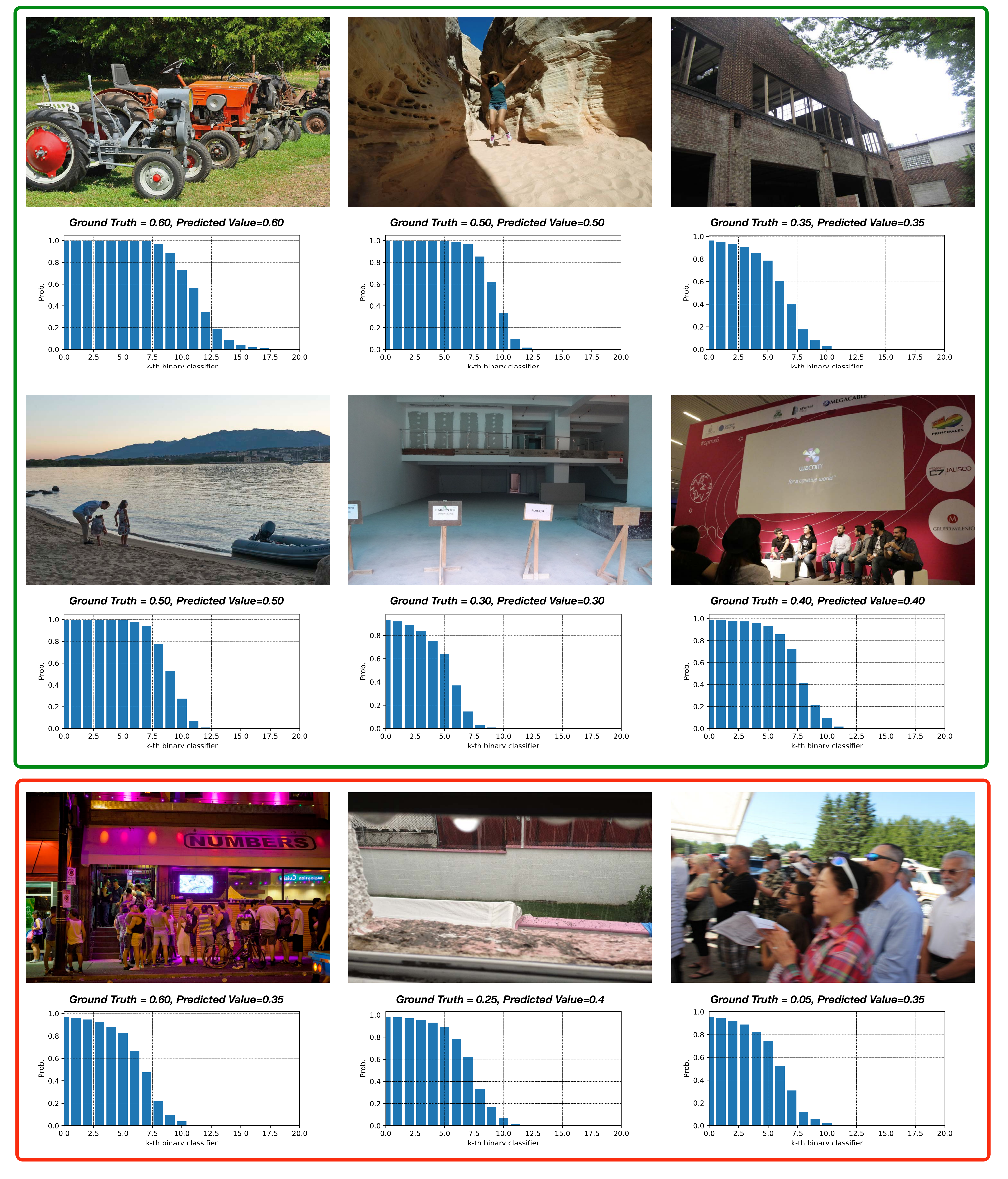}
\caption{The label range of AADB is $[0,1]$ and each label is transformed into a 20 dimensionality ordinal distribution label with a partitioning interval 0.05.}
\label{aesthetic_vis}
\end{figure}

\subsection{Predicted Examples by Our CORF}\label{predict_examples}
Figs.~\ref{age_estimation_vis} and~\ref{aesthetic_vis} show some examples of good and poor results by our CORF on facial age estimation task (MORPH~\cite{ricanek2006morph}) and image aesthetic assessment (AADB~\cite{kong2016photo}) task, respectively. We can see that the proposed approach performs well by estimated the ordinal distribution on age estimation and aesthetic assessment; see the top green boxes of Figs.~\ref{age_estimation_vis} and~\ref{aesthetic_vis}. However, the age estimation accuracy may decrease for the old people because the distribution of the MOPRH dataset is pretty imbalanced and the sample of the old man is pretty small; see the bottom red box of Fig.~\ref{age_estimation_vis}. Besides, the aesthetic assessment accuracy may decrease if the images are blurring; see the bottom red box of Fig.~\ref{aesthetic_vis}.

\subsection{Computational Complexity}
The complexity of our model can be analyzed in two separate parts: feature function and the random forests. When using VGG-16 as the backbone, the computational complexity of the feature function is about 15.3G FLOPs. Let $H$ and $C$ denote the tree depth and the dimension of leaf node, respectively. Then each tree contains $D = 2^{H-1} - 1$ split nodes and $2^{H-1}$ leaf nodes. The complexity of testing a tree is $\mathcal{O}(D\times C)$. On the MORPH dataset that has 43,965 training images 11,100 testing images, our CORF model only takes 6,680s for the training with 27,000 iterations and 17s for testing all 11,100 images, which highlights that our method only takes 1.5ms to process one image. Therefore, our approach can be used in many real-time prediction scenarios.

\section{Conclusion}\label{conc}
In this paper, we proposed a convolutional ordinal regression forest (CORF) to solve the image ordinal estimation by extending the differentiable decision tree to learn an ordinal distribution, which can better preserve the global ordinal relationship and make our model both more stable and accurate. It is the first work to integrate random forest and ordinal regression in a deep learning framework to obtain ordering results for image ordinal estimation tasks. The proposed CORF overcomes many stability issues of previous methods solving a series of binary classifiers independently. As a bonus, feature representation can be learned jointly with random forests in an end-to-end manner. In addition, we provide a detailed derivation for the update rule of the leaf node. Finally, the experimental results on two image ordinal estimation tasks  demonstrate the effectiveness and stability of the proposed CORF.

For future work, it is recommended to apply the CORF to the ordinal ranking in the question answering/information retrieval field with changes in three aspects: 1) task from \emph{ordinal regression} to \emph{ordinal ranking}, network from \emph{convolutional neural network} to \emph{recurrent neural network}, and modality from \emph{image} to \emph{text}.

\IEEEpeerreviewmaketitle

\end{document}